\documentclass{article}

\usepackage{microtype}
\usepackage{graphicx}
\usepackage{subcaption}
\usepackage{booktabs} 
\usepackage{arydshln} 
\usepackage{enumitem}

\usepackage{hyperref}

\usepackage[accepted]{icml2026}

\makeatletter
\renewcommand{\ICML@appearing}{}
\makeatother
\usepackage{amsmath}
\usepackage{amssymb}
\usepackage{mathtools}

\icmltitlerunning{Talker-T2AV: Joint Talking Audio-Video Generation with  Autoregressive Diffusion Modeling}

\begin{document}

\twocolumn[
  \icmltitle{Talker-T2AV: Joint Talking Audio-Video Generation with \\
    Autoregressive Diffusion Modeling}

  \icmlsetsymbol{equal}{*}

  \begin{icmlauthorlist}
    \icmlauthor{Zhen Ye}{equal,HKUST}
    \icmlauthor{Xu Tan}{equal,ind}
    \icmlauthor{Aoxiong Yin}{ZJU}
    \icmlauthor{Hongzhan Lin}{NUS}
    \icmlauthor{Guangyan Zhang}{ind}
    \icmlauthor{Peiwen Sun}{CUHK}
    \icmlauthor{Yiming Li}{HKUST}
    \icmlauthor{Chi-Min Chan}{HKUST}
    \icmlauthor{Wei Ye}{pku}
    \icmlauthor{Shikun Zhang}{pku}
    \icmlauthor{Wei Xue}{HKUST}
  \end{icmlauthorlist}

  \icmlaffiliation{HKUST}{Hong Kong University of Science and Technology}
  \icmlaffiliation{ZJU}{Zhejiang University}
  \icmlaffiliation{NUS}{National University of Singapore}
  \icmlaffiliation{CUHK}{The Chinese University of Hong Kong}
  \icmlaffiliation{pku}{Peking University}
  \icmlaffiliation{ind}{Independent Researcher}

  \icmlcorrespondingauthor{Wei Xue}{weixue@ust.hk}

  \icmlkeywords{talking head generation, joint audio-video generation, autoregressive modeling, diffusion transformer, multimodal generation}

  \vskip 0.3in
]

\printAffiliationsAndNotice{\textsuperscript{*}Equal contribution; the order of the first two authors is arbitrary. }

\begin{abstract}
Joint audio-video generation models, such as dual-branch diffusion
transformers, have demonstrated that generating audio and video within a
unified model yields stronger cross-modal coherence than cascaded
pipelines.  However, these models typically couple the two modalities throughout the entire denoising process via pervasive attention, treating high-level semantics and low-level signal details in a fully entangled manner. We argue that such uniform coupling is suboptimal for talking head synthesis: while audio and facial motion are strongly correlated at the semantic and temporal levels, their low-level realizations---acoustic signals and visual textures---follow distinct rendering processes that do not benefit from continuous entanglement. Joint modeling should therefore occur primarily at the high-level semantic and temporal level, where cross-modal alignment is essential, while low-level refinement is delegated to modality-specific decoders.
Building on this observation, we propose  Talker-T2AV, an autoregressive diffusion framework that realizes this principle. A shared autoregressive language model acts as a high-level temporal planner, jointly reasoning over audio and video in a unified patch-level token space, and two lightweight diffusion transformer heads decode its hidden states into frame-level audio and video latents. Experiments on talking portrait benchmarks show that Talker-T2AV outperforms   dual-branch diffusion transformer baselines in lip-sync accuracy, video quality, and audio quality, and achieves stronger cross-modal consistency than cascaded pipelines. Crucially, operating on short, temporally aligned $1$-D latent sequences also makes generation cheap: Talker-T2AV is more than $50\times$ faster at inference than dual-branch diffusion baselines. The samples, code, and training data are available at \url{https://talker-t2av.github.io/}, \url{https://github.com/zhenye234/Talker-T2AV}, and \url{https://huggingface.co/datasets/HKUSTAudio/Talker-T2AV-Data}, respectively.
\end{abstract}

\section{Introduction}
\label{sec:intro}

Generating synchronized audio and video within a single unified model has emerged as one of the most exciting frontiers in generative AI.
Leading commercial systems such as Sora~2~\citep{openai2025sora2}, Veo~3~\citep{googledeepmind2025veo3}, Kling~3~\citep{kuaishou2026kling3}, and Seedance~2.0~\citep{bytedance2026seedance2} have all adopted joint audio-video generation as a core capability, producing cinematic clips with tightly synchronized dialogue, sound effects, and visual motion.
Rather than producing each modality in isolation, these systems co-generate auditory and visual signals through a unified model, allowing the two modalities to inform each other during synthesis. This tight cross-modal coupling has yielded remarkable advances in temporal synchronization, semantic coherence, and overall perceptual quality compared with earlier cascaded pipelines. Although the details of these commercial systems remain undisclosed, open-source research has converged on the \emph{dual-branch diffusion transformer}~\citep{peebles2023scalable,liu2024syncflow} (dual-DiT): two parallel DiT towers---one for video, one for audio---coupled via bidirectional cross-attention at every block. Representative systems include MOVA, Ovi and LTX-2 \citep{mova2026, ovi2025, hacohen2026ltx}. A consistent finding is that joint modeling outperforms cascaded pipelines in lip-sync accuracy, audio-visual quality, and overall coherence.

However, existing dual-branch diffusion models exhibit three structural limitations when applied to talking head generation.
First, these models couple audio and video through pervasive cross-modal attention at every denoising step, entangling high-level semantic modeling and low-level signal rendering throughout the entire process.
We argue that such uniform coupling is suboptimal for talking heads: while speech and facial motion are strongly correlated at the \emph{semantic} and \emph{temporal} levels, their low-level realizations---acoustic waveforms and visual textures---follow fundamentally distinct rendering processes that do not benefit from continuous cross-modal attention. Second, dual-branch diffusion models are inherently non-autoregressive, and usually commit to a fixed output length before generation begins (e.g., $\sim$5s~\citep{zhang2025uniavgen,ovi2025,wang2025universe,wan2025}).
When the input text exceeds what the predetermined duration can accommodate, the model is forced to compress, truncate, or skip content, severely degrading speech intelligibility.
This fixed-length constraint also precludes adapting to naturally varying speaking rates across languages and speakers.
Third, these models are prohibitively expensive at inference: every denoising step operates on a dense $T\!\times\!H\!\times\!W$ video latent volume comprising thousands to tens of thousands of tokens, and this cost is paid tens of times over the diffusion trajectory.
In practice, released dual-branch systems generate far below real time, which limits their deployment value for interactive talking-head applications.

These limitations suggest a more suitable design principle for talking head generation. Rather than coupling audio and video throughout the entire denoising process, cross-modal interaction should primarily occur at the level of high-level semantic and temporal planning, where speech content and facial dynamics need to be coordinated. Once such shared structure is established, each modality should be rendered through its own specialized generation process to account for the distinct characteristics of acoustic and visual signals. In addition, instead of relying on a fixed-length diffusion procedure, the generation framework should support progressive, context-dependent expansion so that output duration can naturally adapt to the input text and speaking rhythm.
To this end, we propose \textsc{Talker-T2AV}, an autoregressive diffusion framework for joint talking audio-video generation from text. \textsc{Talker-T2AV} factorizes the generation process into two stages: high-level cross-modal modeling in a shared autoregressive backbone, and low-level modality-specific rendering in two independent diffusion transformer heads. We encode speech and video into temporally aligned latent sequences at an identical frame rate and fuse them via element-wise summation at each position in the backbone, which autoregressively predicts the next joint patch conditioned on the text prefix and all preceding context. The two diffusion heads then independently decode each hidden state into frame-level speech and video latents, each specialized for its own signal characteristics. Moreover, the element-wise summation design provides inherent flexibility: when one modality is available as input, its embeddings are directly fed into the backbone while the other is autoregressively predicted, enabling a single model to perform joint audio-video generation, audio-driven talking head synthesis, and video dubbing without any architectural modification or additional fine-tuning.

Contrary to the common expectation that autoregressive decoding is slower than parallel diffusion, this factorization makes \textsc{Talker-T2AV} substantially \emph{faster} than dual-branch diffusion baselines.
Because both modalities are encoded as purely temporal $1$-D latent sequences at $25$\,Hz and further compressed by patching, a five-second clip corresponds to only about $31$ autoregressive steps, and each diffusion head denoises a single four-frame patch instead of a dense spatio-temporal latent volume.
The result is near real-time generation at a fraction of the parameter count, together with streaming output that non-causal diffusion cannot provide.

Our main contributions are summarized as follows:
\textbf{(1)}~We propose Talker-T2AV that decouples joint audio-video generation into high-level cross-modal modeling in a shared autoregressive backbone and low-level rendering in two independent diffusion transformer heads, avoiding the pervasive cross-modal entanglement of dual-branch diffusion transformers while naturally supporting variable-length output.
\textbf{(2)}~We encode audio and video into temporally aligned latent sequences and fuse them via element-wise summation in the backbone. This design enables a single model to perform joint audio-video generation, audio-to-video, and video-to-audio without architectural modification or additional fine-tuning.
\textbf{(3)}~Experiments on talking head benchmarks show that \textsc{Talker-T2AV} outperforms dual-branch diffusion transformer baselines on joint audio-video generation across speech intelligibility, video fidelity, and lip-sync accuracy, and matches or surpasses dedicated systems on audio-driven synthesis and video dubbing.
\textbf{(4)}~Despite being autoregressive, \textsc{Talker-T2AV} is the fastest joint text-to-audio-video system among all compared methods---$24$\,FPS with $1$B parameters versus $\le\!0.41$\,FPS for $7$--$32$B dual-branch diffusion baselines---and additionally supports streaming, low-latency output that non-causal diffusion cannot provide.

\section{Related Work}
\label{sec:related}

\subsection{Joint Audio-Video Generation}

To eliminate the serial audio-then-video dependency, recent work explores joint audio-video generation within a single model.
The dominant paradigm is the \emph{dual-branch diffusion transformer} (dual-DiT), where two parallel DiT towers---one for video, one for audio---are coupled via bidirectional cross-modal attention.
Representative systems include Ovi~\citep{ovi2025}, MOVA~\citep{mova2026}, UniVerse-1~\citep{wang2025universe}, JavisDiT~\citep{liu2025javisdit}, and LTX-2~\citep{hacohen2026ltx};
These general-purpose systems consistently demonstrate that joint modeling yields stronger cross-modal coherence than cascaded pipelines.

Several works apply joint generation specifically to talking portraits.
UniTalking~\citep{li2026unitalking} builds an end-to-end diffusion framework with multi-modal transformer blocks that model fine-grained audio-video temporal correspondence via shared self-attention.
UniAVGen~\citep{zhang2025uniavgen} augments a dual-DiT architecture with a face-aware modulation module to dynamically prioritize salient facial regions. OmniTalker~\citep{wang2025omnitalker} enables a text-driven DiT framework for joint speech and facial video generation with one-shot multimodal style mimicking. Faces that Speak~\citep{jang2024faces} couples TTS and talking face generation via intermediate feature sharing between two separate decoding pipelines, rather than jointly modeling both modalities within a unified generative backbone. AV-Flow~\citep{chatziagapi2025av} also uses dual diffusion transformers through intermediate highway layers for joint audio-visual generation of 4D talking avatars.

Despite improved coherence, existing joint frameworks exhibit three structural limitations for talking head synthesis.
First, pervasive cross-modal attention couples high-level semantic modeling with low-level signal rendering throughout the entire denoising process; we argue this uniform coupling is suboptimal \emph{in this setting}, because while speech audio and talking head video share strong semantic and temporal correlations, their low-level realizations follow distinct rendering processes that do not benefit from continuous entanglement.
Second, these models are inherently non-causal: they denoise the full sequence jointly, precluding streaming output and variable-length generation, while essential for practical deployment.
Third, they are computationally heavy at inference, because each denoising step attends over a dense spatio-temporal video latent volume; as we show in \S\ref{sec:efficiency}, released dual-DiT systems generate two to three orders of magnitude below real time.

\subsection{Audio-to-Video Generation}

Audio-driven talking head generation assumes that a speech signal is already available and focuses on synthesizing synchronized facial video from it.
Early methods such as Wav2Lip~\citep{prajwal2020lip} introduced a lip-sync expert discriminator to enforce speech--lip alignment, while SadTalker~\citep{zhang2023sadtalker} extended the scope from lip-only synchronization to realistic head pose and expression via 3D motion coefficients.
Subsequent work pursued higher visual fidelity through diverse technical routes:
GeneFace++~\citep{ye2023geneface++} and Real3D-Portrait~\citep{ye2024real3d} leveraged neural radiance fields for 3D-consistent rendering;
EMO~\citep{tian2024emo} adopted audio-to-video diffusion to bypass explicit structural intermediates;
Hallo~\citep{xu2024hallo,cui2024hallo2,cui2025hallo3} pushed diffusion-based portrait animation toward longer duration and higher resolution;
and EchoMimic~\citep{chen2025echomimic} introduced editable landmark conditioning for finer-grained control.
More recently, FLOAT~\citep{ki2025float} and Ditto~\citep{li2025ditto} brought flow matching and explicit motion latent spaces into the pipeline, improving temporal consistency and sampling efficiency.
Diffusion-based methods have become the dominant paradigm in this line of work, largely because the conditioning audio already specifies the output length, speaking rate, and prosodic structure, leaving the model only responsible for rendering the corresponding visual signal within a known temporal information.
Compared with joint audio-video modeling, however, cascaded pipelines assume audio is already available, either from ground-truth recordings or a separate TTS system, and model video generation conditionally on it, limiting cross-modal coherence. When both audio and video must be jointly generated from text, none of this temporal information is available as input, and duration, rhythm, and prosody must all be inferred by the model itself, posing additional challenges for fixed-length generation paradigms such as diffusion models.

\section{Method}
\label{sec:method}

\begin{figure*}[t]
  \centering
\includegraphics[width=0.99\textwidth]{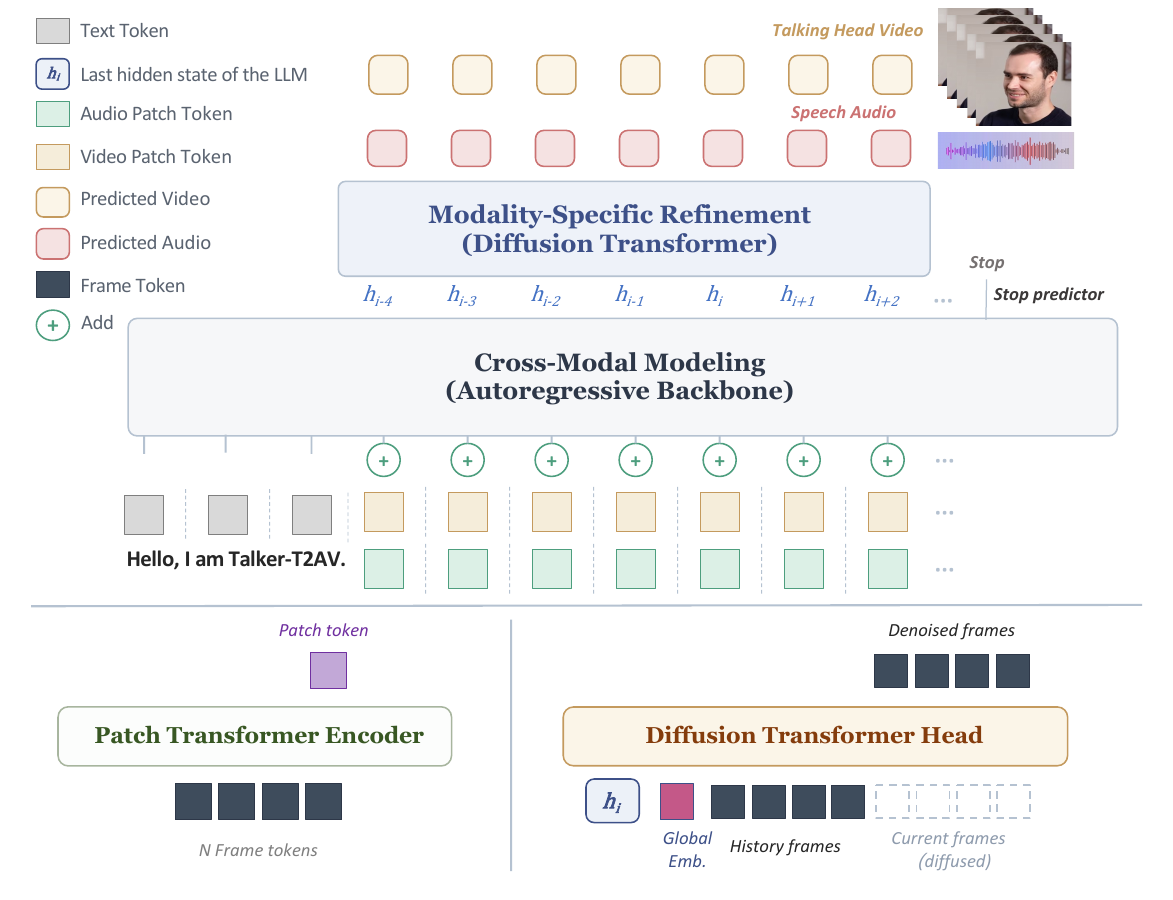}
   \caption{\textbf{Overview of \textsc{Talker-T2AV}.}
    \textbf{Top:} The autoregressive backbone processes a unified causal
    sequence in which the full text token sequence appears first as a prefix,
    followed by joint audio-video patches where audio and video patch
    embeddings are element-wise summed at each position.
    The backbone hidden states are then decoded by two
    modality-specific diffusion transformer heads into audio waveforms
    and portrait video, respectively.
    \textbf{Bottom right:} Each diffusion transformer head takes the hidden
    state $\mathbf{h}_i$, a global embedding, and a context window of
    history frames as conditioning, and denoises the current frame via
    flow matching.
    \textbf{Bottom left:} The patch transformer encoder compresses $N$
     audio/video frames into a single patch token, reducing sequence length for efficient autoregressive modeling.}
  \label{fig:overview}
\end{figure*}

\subsection{Overview}
The overall architecture of \textsc{Talker-T2AV} is illustrated in Figure~\ref{fig:overview}.
Our framework takes as input a text transcript along with a reference speech and an identity image jointly generates an audio latent sequence~$\mathbf{A}=(\mathbf{a}_1,\dots,\mathbf{a}_N)$ and a talking head video latent sequence~$\mathbf{V}=(\mathbf{v}_1,\dots,\mathbf{v}_N)$, where each~$\mathbf{a}_i\!\in\!\mathbb{R}^{P\times d_a}$ and~$\mathbf{v}_i\!\in\!\mathbb{R}^{P\times d_v}$ denote a patch of~$P$ consecutive frames, and $N$ is determined by a learned stop predictor.

The input to the autoregressive backbone is organized as a  text-driven  causal sequence:
the full text token sequence $\mathbf{t}=(t_1,\dots,t_M)$ is placed at the beginning as a prefix,
followed by $N$ joint audio-video patch tokens in which the audio and video patch embeddings are element-wise summed at each position.

We model the joint distribution autoregressively at the patch level:
\begin{equation}
  p(\mathbf{A},\mathbf{V}\mid\mathbf{t})
  \;=\;\prod_{i=1}^{N}\,
  p\!\bigl(\mathbf{a}_i,\mathbf{v}_i
    \mid \mathbf{a}_{<i},\,\mathbf{v}_{<i},\,\mathbf{t}\bigr),
  \label{eq:ar-joint}
\end{equation}
where $\mathbf{t}$ denotes the full text prefix that conditions all patch positions uniformly.
Each step is factorized into two stages following a coarse-to-fine paradigm.

\subsection{Stage 1: Cross-Modal Modeling}

The first stage is carried out by a shared causal language model that
jointly reasons over text, speech, and video in a unified sequence.
We describe its three key components below.

\paragraph{Temporally Aligned Cross-Modal Features.}
Speech and video are fundamentally different modalities.
In typical video generation pipelines, a video VAE encodes each frame
into a 2-D spatial grid, yielding a $T\!\times\!H\!\times\!W$ token
volume, whereas speech is naturally a 1-D temporal sequence.
This structural mismatch means there is no straightforward way to
align the two representations frame by frame;
prior joint models such as dual-DiT~\citep{mova2026,ovi2025} therefore rely on
cross-attention layers to learn the correspondence, which adds
considerable optimization difficulty.

We argue that audio and  video are \emph{inherently} aligned
along the time axis---the articulation at time~$t$ directly determines
the speech content, lip shape, and facial expression at the same instant.
To exploit this natural alignment, we deliberately choose feature
extractors that produce \emph{purely temporal} sequences at an
identical frame rate for both modalities, eliminating the need for any
cross-attention or learned temporal alignment module.

Specifically, for the video modality, we adopt
LIA-X~\citep{wang2025lia,wang2022latent} to obtain a compact motion latent space that represents each
frame as a single vector
$\hat{\mathbf{v}}_t\!\in\!\mathbb{R}^{d_v}$ at 25\,Hz
(details in Appendix~\ref{app:liax}).
For the audio modality, we employ WhisperX-VAE~\citep{radford2023robust,ye2025codec} to encode the waveform
into a continuous latent
$\hat{\mathbf{a}}_t\!\in\!\mathbb{R}^{d_a}$ at the same 25\,Hz rate
(details in Appendix~\ref{app:whispervae}).
Both encoders are frozen during training.
Because the two streams share an identical temporal resolution, the
$t$-th speech frame and the $t$-th motion frame correspond to the same
physical instant, yielding temporally aligned cross-modal features by
construction.

\paragraph{Autoregressive Backbone.}
The autoregressive backbone forms the Cross-Modal Modeling
stage of our hierarchical architecture.
It is responsible for capturing high-level cross-modal temporal
structure and producing shared hidden states that condition modality-specific decoding.

At each position~$i$, the audio patch embedding
$\mathbf{e}^a_i\!\in\!\mathbb{R}^{D}$ and the video patch embedding
$\mathbf{e}^v_i\!\in\!\mathbb{R}^{D}$ are element-wise summed to
form a joint input token.
Together with the text prefix, these joint tokens are fed into a
 causal language model, which outputs a shared
hidden state:
\begin{equation}
  \mathbf{h}_i = \mathrm{LM}\!\bigl(
    \mathbf{e}^a_{\le i}+\mathbf{e}^v_{\le i},\;\mathbf{t}\bigr),
  \label{eq:backbone}
\end{equation}
where $\mathbf{t}$ is the full text prefix and
$\mathbf{e}^a_{\le i}+\mathbf{e}^v_{\le i}$ denotes the sequence of
element-wise-summed patch tokens up to position~$i$.
The same hidden state~$\mathbf{h}_i$ is then dispatched to both the
audio and video diffusion transformer heads as conditioning
(\S\ref{sec:stage2}), so that high-level cross-modal coherence is
captured in a shared representation space while low-level
rendering is left to modality-specific decoders.

We note that element-wise summation is not the only possible design
for fusing the two modalities.
A natural alternative is \emph{interleaving}---placing speech and
video tokens as separate entries in an alternating sequence.
Another option, inspired by traditional cascade talking-head pipelines, is to place audio tokens several frames ahead of their corresponding video tokens.
We compare these strategies empirically in \S\ref{sec:ablation} and
find that the summation design achieves the best performance of
synchronization quality and training efficiency.

Moreover, the element-wise summation design endows the model with
inherent flexibility for \emph{uni-modal conditioned generation} at
inference time.
When one modality is available as input, its ground-truth patch embeddings are summed with the autoregressively predicted embeddings of the other modality at each position before being fed into the autoregressive backbone.
This enables two additional applications without any architectural
modification or fine-tuning:
(1)~\emph{video dubbing}---given a silent talking-head video and a
text transcript, the model generates synchronized speech that matches
both the linguistic content and the visual articulation;
(2)~\emph{audio-driven talking head generation}---given an audio signal and a reference face image, the model generates lip-synced video by
conditioning on the ground truth speech embeddings.
In both cases, the shared backbone leverages the available modality to
inform the hidden states, and only the corresponding diffusion head
is activated for decoding.
This unified formulation contrasts with prior cascade approaches that
require entirely separate models for each modality generation direction.

\paragraph{Patch Transformer Encoder.}

Operating directly on the frame level would produce
prohibitively long sequences for the autoregressive backbone, resulting in quadratic
attention cost and slow inference.
Therefore, we group every $P$ consecutive frames into a
single patch and compress each patch into one token, reducing
the sequence length by a factor of~$P$ and accelerating both training
and inference accordingly.

Concretely, each modality has a dedicated Patch Transformer Encoder.
It first projects the $P$ frame-level latents into the backbone
hidden dimension~$D$ via a linear layer, prepends a learnable
\texttt{[CLS]} token, and processes the resulting $(P\!+\!1)$-length
sequence with a small bidirectional Transformer.
The output at the \texttt{[CLS]} position serves as the compressed
patch representation.
The speech and video patch embeddings are then element-wise summed at
each position to form the joint patch token fed into the backbone
(Eq.~\eqref{eq:backbone}).

\paragraph{Stop Predictor.}

Since the output length is not known in advance, we attach an
 MLP stop predictor to the last LM backbone hidden state:
$p_{\mathrm{stop}}(i)=\mathrm{sigmoid}\!\bigl( \mathrm{MLP}(\mathbf{h}_i)\bigr)$.
It is trained with binary cross-entropy loss, where the positive class weight is set to
the ratio of continue labels to stop labels in each batch to
compensate for the severe class imbalance.
 This mechanism enables our model to produce outputs of arbitrary
length, adapting naturally to varying text inputs and speaking rates---a
capability that pure DiT architectures require additional mechanisms to support.

\subsection{Stage 2: Modality-Specific Refinement}
\label{sec:stage2}

As shown in Figure~\ref{fig:overview}, the shared hidden state~$\mathbf{h}_i$ produced by the autoregressive
backbone encodes high-level cross-modal information---what
phoneme to articulate, which expression to display, and how the two
modalities coordinate at position~$i$.
However, speech and video latent spaces differ fundamentally in
dimensionality, signal dynamics, and noise characteristics
($d_a$-dimensional acoustic features versus $d_v$-dimensional facial
motion coefficients).
Forcing a single decoder to reconstruct both modalities would conflate
these disparate low-level properties.
We therefore adopt a \emph{modality-specific decoding} design: two
independent diffusion heads, each a lightweight Diffusion Transformer
(DiT), take the same hidden state~$\mathbf{h}_i$ as semantic
conditioning and decode it into the corresponding patch of~$P$
continuous latent frames.

Each diffusion head is a bidirectional Transformer whose input sequence
is formed by concatenating four components:
(i)~the hidden state~$\mathbf{h}_i$ added with a sinusoidal timestep
embedding of the diffusion step~$\tau$, serving as the semantic anchor;
(ii)~a global condition vector that supplies identity information
(\emph{e.g.}, a speaker embedding for speech~\citep{Qwen3-TTS,zhang2025minimax}, or the first-frame motion
vector for video~\citep{ki2025float});
(iii)~a context window consisting of the latent frames from the
immediately preceding patch, providing short-term continuity cues; and
(iv)~the noisy target~$\mathbf{x}_\tau$, which the model learns to
denoise.
The Transformer processes this composite sequence with full
bidirectional attention, and the output at the positions corresponding
to~$\mathbf{x}_\tau$ is projected back to the latent dimension as the
predicted velocity field.

The audio head and the video head share the same architecture
  but maintain entirely separate parameters, allowing each
head to specialize in the statistics of its respective latent space.
We train both heads with the optimal-transport conditional flow matching
(OT-CFM) objective~\citep{lipman2022flow,tong2023improving}.
Given a clean latent patch
$\mathbf{x}_0\!\in\!\mathbb{R}^{d\times P}$,
we construct a noisy sample
$\mathbf{x}_\tau=(1-\tau)\,\mathbf{x}_0+\tau\,\mathbf{z}$
with $\mathbf{z}\!\sim\!\mathcal{N}(\mathbf{0},\mathbf{I})$ and
$\tau$ drawn from a logit-normal distribution.
The head is trained to predict the velocity
$\mathbf{v}=\mathbf{z}-\mathbf{x}_0$:
\begin{equation}
  \mathcal{L}_{\mathrm{cfm}}
  \;=\;
  \mathbb{E}_{\tau,\mathbf{z}}\!\Bigl[
    \bigl\lVert
      v_\theta\!\bigl(\mathbf{x}_\tau,\,\tau,\,\mathbf{h}_i,\,\mathbf{c}\bigr)
      -\mathbf{v}
    \bigr\rVert^2
  \Bigr],
  \label{eq:cfm}
\end{equation}
where $\mathbf{c}$ denotes the concatenation of global condition and
context.
During training, the backbone hidden state~$\mathbf{h}_i$ is
randomly dropped with a small probability to enable classifier-free
guidance~(CFG)~\citep{ho2022classifier} at inference time.

 \subsection{Multi-Task Training Objective}
High-quality audio-video paired data is considerably scarcer than
audio-only corpora.
Training exclusively on paired data limits the modeling capacity, leading to degraded speech intelligibility.
Our architecture naturally supports a mixed-task training regime:
each sample carries a learnable task tag embedding
$\mathbf{e}_{\mathrm{task}}\!\in\!\{\mathbf{e}_{\mathrm{TTS}},\,
\mathbf{e}_{\mathrm{T2AV}}\}$ placed to the first token,
and for TTS-only samples the motion branch input is replaced by a learnable padding embedding~$\mathbf{e}_{\mathrm{pad}}$ with the motion loss zeroed out.

The full training loss is:
\begin{equation}
  \mathcal{L}
  \;=\;
  \mathcal{L}_{\mathrm{cfm}}^{\mathrm{audio}}
  + \lambda\,\mathcal{L}_{\mathrm{cfm}}^{\mathrm{video}}
  + \alpha\,\mathcal{L}_{\mathrm{stop}},
  \label{eq:total}
\end{equation}
where $\lambda$ and $\alpha$ are scalar weights.
All components are trained jointly from scratch end-to-end.
By mixing in large-scale TTS data, the shared backbone and the speech
diffusion head are exposed to far more diverse text--pronunciation
pairs, learning a more robust mapping from text to linguistic content.
We find this substantially lowers the word error rate of generated
speech while also improving lip articulation accuracy, as the
backbone's stronger text-to-pronunciation modeling provides more
precise phonetic cues for the video diffusion head.

\section{Experiments}
\label{sec:experiment}

\subsection{Implementation Details}
\label{ssec:Implementation}

Unlike dual-branch diffusion transformer approaches that require multi-stage training, separately pre-training the audio branch, the video branch, and then jointly fine-tuning~\citep{ovi2025,mova2026,hacohen2026ltx}, \textsc{Talker-T2AV} is trained in a single stage with multi-task learning.
We collect a dataset of approximately 1M talking-head audio-video clips with aligned transcripts from publicly available online sources, applying a multi-stage filtering pipeline that includes face detection, quality scoring, and transcription, for the joint text-to-audio-video (T2AV) task. For the text-to-speech (TTS) task, we use the same Emilia dataset~\citep{he2024emilia} as UniAVGen~\citep{zhang2025uniavgen} used.
During training, each mini-batch is composed of equal parts T2AV and TTS samples, governed by the task tag mechanism described in \S\ref{sec:stage2}.
 The autoregressive backbone is initialized from Qwen3-0.6B~\citep{yang2025qwen3}.
Each modality-specific diffusion head is an 8-layer bidirectional Transformer with 8 attention heads and a hidden dimension of 1024.
Each Patch Transformer Encoder consists of 4 bidirectional Transformer layers with 8 attention heads and a hidden dimension of 1024.
 The patch size is $P\!=\!4$ frames. Each diffusion transformer head predicts a patch of $P\!=\!4$ latent frames and is conditioned on a context window of $P\!=\!4$ frames, providing short-term continuity cues for smooth inter-patch transitions. We train the full model end-to-end for 200{,}000 steps with a global batch size of 256 using AdamW~\citep{loshchilov2017decoupled} in bfloat16 mixed precision.
The learning rate is set to $1\!\times\!10^{-4}$ with a linear warm-up over the first 3\% of training steps.
In the total loss (Eq.~\eqref{eq:total}), the video diffusion loss weight is $\lambda\!=\!8$ and the stop loss weight is $\alpha\!=\!1$.

 \subsection{Evaluation Setup}

During inference, the overall pipeline proceeds as follows.
The input text is first tokenized and placed as a prefix to the autoregressive backbone.
At each autoregressive step, the backbone produces a shared hidden state~$\mathbf{h}_i$, which is fed into the audio and video diffusion transformer heads.
Each head independently denoises a patch of $P$ latent frames via flow matching solved with an Euler ODE sampler using 10 steps, a temperature of $t{=}0.7$, and a classifier-free guidance scale of $\mathrm{CFG}{=}2.0$.
Generation terminates when the stop predictor outputs a probability exceeding $0.5$.
The resulting audio latent sequence is decoded back to a waveform by the audio VAE decoder, and the video motion latent sequence is decoded to portrait video by the LIA-X decoder
(see Appendices~\ref{app:liax} and~\ref{app:whispervae} for details of these autoencoders).

\subsection{Comparisons with Joint Audio-Video Generation}

\begin{table*}[t]
\caption{Comparison with joint text-to-audio-video generation methods on both test sets.
Methods above the dashed line target \emph{general} audio-video generation, while UniAVGen and our method focus on facial talking-head video generation.
Best results are in \textbf{bold}, second best are \underline{underlined}.}
\label{tab:joint-compare}
\centering
\resizebox{\textwidth}{!}{%
\begin{tabular}{l cc cc cc cc cc cc}
\toprule
& \multicolumn{6}{c}{\textbf{Chinese}}
& \multicolumn{6}{c}{\textbf{English}} \\
\cmidrule(lr){2-7} \cmidrule(lr){8-13}
& \multicolumn{2}{c}{Audio} & \multicolumn{2}{c}{Video} & \multicolumn{2}{c}{Sync}
& \multicolumn{2}{c}{Audio} & \multicolumn{2}{c}{Video} & \multicolumn{2}{c}{Sync} \\
\cmidrule(lr){2-3} \cmidrule(lr){4-5} \cmidrule(lr){6-7}
\cmidrule(lr){8-9} \cmidrule(lr){10-11} \cmidrule(lr){12-13}
\textbf{Method}
& CER$\downarrow$ & UTMOS$\uparrow$
& FID$\downarrow$ & FVD$\downarrow$
& C$\uparrow$ & D$\downarrow$
& WER$\downarrow$ & UTMOS$\uparrow$
& FID$\downarrow$ & FVD$\downarrow$
& C$\uparrow$ & D$\downarrow$ \\
\midrule
MoVA~\citep{mova2026}
& 0.359 & 1.979 & 38.87 & 249.20 & 3.008 & 10.719
& 0.317 & 3.033 & 34.75 & 301.82 & 2.982 & 11.107 \\
Ovi~\citep{ovi2025}
& 0.873 & 2.085 & 29.75 & 224.28 & 1.496 &  11.515
& 0.296 & 3.030 & 33.84 & 284.56 & 4.166 & \underline{9.582} \\
LTX-2~\citep{hacohen2026ltx}
& 0.461 & 2.053 & 32.49 & 318.13 &1.656 & 12.387
& \underline{0.257} & 2.769 & \underline{27.46} & \underline{272.78} & \underline{4.671} & 9.642 \\
UniVerse-1~\citep{wang2025universe}
& 0.715 & 1.511 & 19.49 & 237.41 & 0.661 & 13.678
& 0.385 & 1.690 & 36.50 & 409.58 & 1.092 & 13.906 \\
\hdashline
\noalign{\vskip 2pt}
UniAVGen~\citep{zhang2025uniavgen}
& \underline{0.265} & \textbf{2.197} & \textbf{15.30} & \underline{157.92} &  \underline{3.168} & \underline{9.956}
& 0.302 & \textbf{3.459} & 35.27 & 298.27 & 2.555 & 11.378 \\
\midrule
\textbf{Ours}
& \textbf{0.148} & \underline{2.136} & \underline{17.63} & \textbf{103.31} & \textbf{5.470} & \textbf{8.793}
& \textbf{0.055} & \underline{3.458} & \textbf{24.32} & \textbf{246.39} & \textbf{6.330} & \textbf{8.505} \\
\bottomrule
\end{tabular}%
}
\end{table*}

\paragraph{Evaluation Metrics.}
We evaluate generated results along three dimensions: \textit{audio quality}, \textit{video quality}, and \textit{audio-visual synchronization}.
For audio quality, we report Character Error Rate (CER) for Chinese and Word Error Rate (WER) for English, computed by transcribing the generated speech with Qwen3-ASR~\citep{shi2026qwen3}; and UTMOS~\citep{saeki2022utmos} for speech naturalness.
For video quality, we report Fr\'{e}chet Inception Distance (FID)~\citep{heusel2017gans} and Fr\'{e}chet Video Distance (FVD)~\citep{unterthiner2018towards} to measure the distributional fidelity of generated frames and videos, respectively.
For audio-visual synchronization, we adopt the SyncNet~\citep{chung2016out} evaluation protocol and report Confidence (C) and minimum Distance (D), where higher C and lower D indicate tighter lip-audio alignment.

\paragraph{Evaluation Data.}
\label{ssec:setup}
We construct two test sets covering Chinese and English textual transcription, respectively, each containing 200 videos.
The Chinese test set is sampled from DH-FaceVid-1K~\citep{di2025dh}, a large-scale, high-quality Chinese-dominant face video dataset.
The English test set is composed of clips drawn from HDTF~\citep{zhang2021flow} and the Hallo3~\citep{cui2025hallo3}.
All test clips are paired with ground-truth text transcripts, audio, and video. Both the text-to-audio-video (T2AV) and audio-to-video (A2V) evaluations use the same test sets.

\paragraph{Compared Methods.}
We compare with five representative joint audio-video generation systems, all built upon the dual-branch diffusion transformer (dual-DiT) paradigm: two parallel DiT towers for audio and video coupled via bidirectional cross-modal attention throughout the denoising process.
Among them, four target \emph{general-purpose} audio-video synthesis:
\textbf{MoVA}~\citep{mova2026}, which focuses on scalable synchronized audio-video generation;
\textbf{Ovi}~\citep{ovi2025}, which augments the dual-DiT with multi-modal alignment for general audio-video generation;
\textbf{LTX-2}~\citep{hacohen2026ltx}, a representative open-source joint audio-video framework;
and \textbf{UniVerse-1}~\citep{wang2025universe}, a unified dual-DiT system for joint audio-video synthesis.
In contrast, \textbf{UniAVGen}~\citep{zhang2025uniavgen} specifically focuses on talking-head generation and extends the dual-DiT architecture with face-aware dynamic modulation---the same application domain as our method.
All baselines are non-causal and generate audio and video of a fixed, predetermined length in a single forward pass, whereas our method produces output autoregressively.

\paragraph{Results.}
Table~\ref{tab:joint-compare} compares our method with all five baselines on both test sets.
For audio quality, our decoupled architecture yields substantially better speech content accuracy and naturalness, achieving the lowest CER on Chinese and the lowest WER on English.
This advantage stems from two complementary design choices.
First, separating modality-specific generation into independent diffusion heads
prevents the audio branch from being distorted by entangled video features,
allowing it to focus on faithful linguistic content reproduction.
Second, all dual-DiT baselines produce audio and video of a fixed
predetermined length. When the input text exceeds this budget, the model must either accelerate
the speaking rate, truncate the utterance, or skip content, all of which
severely degrade intelligibility.
Our autoregressive design, in contrast, produces output of variable length that naturally adapts to the text, eliminating this length mismatch. For video quality, our method attains the best FVD on both test sets and the best or second-best FID, with particularly large gains over general-purpose baselines that do not model facial structure explicitly. Compared with UniAVGen, the strongest facial-focused baseline, we substantially reduce FVD on both Chinese and English test sets, demonstrating that the semantic planning stage of our AR backbone provides rich cross-modal conditioning that directly benefits visual fidelity. The resulting videos also exhibit smooth facial motion trajectories, reflecting the temporal coherence captured in the autoregressive token sequence. As for audio-visual synchronization, our model achieves the highest SyncNet Confidence and the lowest SyncNet Distance on both test sets, outperforming all baselines. This result supports our core hypothesis in the talking-head setting: given frame-aligned audio and motion latents, the autoregressive backbone learns cross-modal alignment through joint semantic planning, and the modality-specific diffusion heads can faithfully render synchronized outputs without requiring additional explicit synchronization losses or cross-attention at the decoding stage.

\begin{table}[t]
\caption{Inference efficiency on a single NVIDIA H20 GPU, measured as generated video frames per second of wall-clock time (FPS) on 5\,s clips.
Since the output video runs at 25\,Hz, FPS${>}25$ corresponds to faster than real time.
Our audio-to-video parameter count is lower because, with audio given, only the video diffusion head runs while the audio head is skipped.
Audio-to-video baselines are detailed in \S\ref{sec:a2v}.}
\label{tab:efficiency}
\centering
\resizebox{\columnwidth}{!}{%
\begin{tabular}{l cc | l cc}
\toprule
\multicolumn{3}{c}{\textbf{Text-to-Audio-Video}}
& \multicolumn{3}{c}{\textbf{Audio-to-Video}} \\
\cmidrule(lr){1-3} \cmidrule(lr){4-6}
\textbf{Method} & \textbf{Params} & \textbf{FPS}$\uparrow$
& \textbf{Method} & \textbf{Params} & \textbf{FPS}$\uparrow$ \\
\midrule
UniVerse-1 & 7.1B  & 0.41 & FLOAT     & 0.3B & 39 \\
UniAVGen   & 7.1B  & 0.32 & Ditto     & 0.2B & 47 \\
Ovi        & 10.9B & 0.20 & EchoMimic & 1.3B & 0.98 \\
LTX-2      & 19B   & 0.17 & Sonic     & 1.5B & 0.77 \\
MoVA       & 32B   & 0.02 & --        & --   & -- \\
\midrule
\textbf{Ours} & \textbf{1B} & \textbf{24}
& \textbf{Ours} & 0.8B & 30 \\
\bottomrule
\end{tabular}%
}
\end{table}

\subsection{Inference Efficiency}
\label{sec:efficiency}

Autoregressive decoding is commonly assumed to be slower than parallel diffusion, since output is produced step by step rather than in a single forward pass.
Table~\ref{tab:efficiency} shows that the opposite holds here: \textsc{Talker-T2AV} is the fastest joint text-to-audio-video system among all compared methods while also being the smallest.
At 24\,FPS it is roughly $59\times$ faster than UniVerse-1, $75\times$ faster than the domain-matched UniAVGen, and three orders of magnitude faster than MoVA, using 1B parameters against their 7--32B.
Since the generated video runs at 25\,Hz, this corresponds to essentially real-time synthesis, whereas every dual-DiT baseline needs minutes of computation per second of output.

The gap follows directly from our representation and factorization rather than from implementation tuning.
Both modalities are encoded as purely temporal 1-D latents at 25\,Hz and compressed by a factor of $P\!=\!4$ through patching, so a 5\,s clip is only $31.25$ autoregressive patch steps and the full backbone sequence stays under 50 tokens including the text prefix.
Each diffusion head then denoises one four-frame patch, an attention window of fewer than ten positions.
Dual-DiT systems instead denoise a dense $T\!\times\!H\!\times\!W$ video latent volume of thousands to tens of thousands of tokens, and pay this quadratic-attention cost at every step of the diffusion trajectory, which dominates their runtime.
Confining cross-modal interaction to a short high-level token sequence is therefore what makes joint modeling cheap, not merely accurate.

The same property holds in the audio-driven setting: at 30\,FPS \textsc{Talker-T2AV} runs faster than real time and is competitive with dedicated real-time renderers such as FLOAT and Ditto, while remaining more than an order of magnitude faster than diffusion-based portrait animators like EchoMimic and Sonic.
Beyond raw throughput, the causal backbone emits patches incrementally, so the first audio-video chunk is available after a single autoregressive step.
Non-causal dual-DiT models must complete the entire denoising trajectory for the whole clip before any output can be decoded, which rules out streaming regardless of their total throughput---a practical advantage for interactive talking-head deployment.

 \begin{table}[t]
\caption{Audio-driven talking head comparison. Each cell shows Chinese\,/\,English results.}
\label{tab:a2v}
\centering
\resizebox{\columnwidth}{!}{%
\begin{tabular}{l c c c c}
\toprule
\textbf{Method} & \textbf{FID}$\downarrow$ & \textbf{FVD}$\downarrow$ & \textbf{Sync-C}$\uparrow$ & \textbf{Sync-D}$\downarrow$ \\
\midrule
FLOAT~\citep{ki2025float}
  & 29.71\,/\,32.24 & 222.52\,/\,360.68 & \underline{2.96}\,/\,3.21 & \underline{10.11}\,/\,10.28 \\
EchoMimic~\citep{chen2025echomimic}
  & 33.43\,/\,42.65 & 273.65\,/\,513.64 & 2.19\,/\,3.41 & 10.88\,/\,10.23 \\
Sonic~\citep{ji2025sonic}
  & \textbf{16.17}\,/\,\underline{24.51} & \textbf{106.57}\,/\,\underline{284.61} & 1.85\,/\,\underline{5.34} & 11.36\,/\,\textbf{8.70} \\
Ditto~\citep{li2025ditto}
  & 17.98\,/\,28.73 & 187.54\,/\,304.72 & 1.77\,/\,4.24 & 11.81\,/\,10.04 \\
AniPortrait~\citep{wei2024aniportrait}
  & 23.63\,/\,29.65 & 336.80\,/\,453.08 & 1.14\,/\,2.59 & 12.42\,/\,11.38 \\
\midrule
\textbf{Ours}
  & \underline{17.32}\,/\,\textbf{24.46} & \underline{107.09}\,/\,\textbf{243.51} & \textbf{3.97}\,/\,\textbf{5.85} & \textbf{10.09}\,/\,\underline{9.03} \\
\bottomrule
\end{tabular}%
}
\end{table}

\subsection{Comparisons on Audio-to-Video Generation}
\label{sec:a2v}

As discussed in \S\ref{sec:method}, the element-wise summation design of our autoregressive backbone naturally supports uni-modal conditioned generation.
By feeding patch embeddings from ground-truth audio into the backbone while autoregressively predicting only the video stream, our model can produce lip-synced talking head video from a given audio signal without any architectural modification or additional fine-tuning.
Although audio-driven generation is not the primary objective of our framework, we evaluate it here to demonstrate the generality of the model.

\paragraph{Evaluation Metrics.}
We adopt the same video quality and synchronization metrics as in \S\ref{ssec:setup}: FID, FVD, and SyncNet Confidence\,/\,Distance (C\,/\,D). Since the driving audio is given rather than generated, audio-side metrics are not applicable in this setting.

\paragraph{Compared Methods.}
We compare with five widely used audio-driven talking head methods:
\textbf{FLOAT}~\citep{ki2025float} introduces flow matching into a learned motion latent space for efficient and temporally consistent portrait animation;
\textbf{EchoMimic}~\citep{chen2025echomimic} combines audio conditioning with editable landmark inputs for fine-grained facial control;
\textbf{Sonic}~\citep{ji2025sonic} is a diffusion-based audio-driven portrait system that achieves strong lip-sync, particularly on English data;
\textbf{Ditto}~\citep{li2025ditto} builds an identity-agnostic motion space coupled with a neural renderer for real-time talking head synthesis;
and \textbf{AniPortrait}~\citep{wei2024aniportrait} generates portrait video from audio via intermediate facial landmarks.
All baselines are dedicated audio-to-video systems that take a pre-recorded or TTS-generated audio waveform as input.
For a fair comparison, we provide the same ground-truth audio to both our model and the baselines, so that any difference in video quality and lip-sync accuracy reflects the generation model itself rather than upstream speech quality.

\paragraph{Results.}

Table~\ref{tab:a2v} compares our model with five dedicated audio-driven talking head baselines on both the Chinese and English test sets.
Despite not being specifically designed or optimized for this setting, \textsc{Talker-T2AV} achieves highly competitive results across all metrics.
On video fidelity, our model ranks first or second in both FID and FVD on each test set, closely matching or surpassing the best dedicated baseline.
On lip-sync accuracy, our model consistently achieves the best or second-best SyncNet Confidence and Distance across both languages, with particularly strong Sync-C scores that rank first on both test sets.
Taking all metrics together, no single baseline outperforms our model comprehensively: our approach delivers the strongest overall balance of video quality and lip-sync accuracy.
This suggests that the cross-modal modeling learned during joint text-to-audio-video training transfers effectively to the audio-conditioned setting---the backbone has internalized the temporal correspondence between speech and facial motion, enabling precise lip articulation even when speech is provided as input rather than jointly generated.
Notably, all compared methods are single-purpose audio-to-video systems, whereas our single unified model supports joint text-to-audio-video generation, audio-driven talking head synthesis, and video dubbing within the same architecture without any modification or additional fine-tuning.

\subsection{Comparisons on Video-to-Audio Dubbing}
\begin{table}[t]
\centering
\caption{Comparison on the Chem  benchmark for video dubbing.}
\label{tab:chem2v2c}
\resizebox{\linewidth}{!}{%
\begin{tabular}{l|cccc}
\toprule
\textbf{Methods} & \textbf{DD}$\downarrow$ & \textbf{EMO-SIM (\%)}$\uparrow$ & \textbf{WER (\%)}$\downarrow$ & \textbf{UTMOS}$\uparrow$ \\
\midrule
Speak2Dub~\citep{zhang2024speaker}
  & 0.5873 & 59.72 & 23.78 & 2.74 \\
StyleDubber~\citep{cong2024styledubber}
  & 0.5627 & 58.54 & 25.43 & 1.95 \\
DeepDubber~\citep{zheng2025deepdubber}
  & 0.5756 & 56.42 & 35.88 & 2.03 \\
ProDubber~\citep{zhang2025prosody}
  & 0.5650 & 65.98 & 14.33 & 2.91 \\
InstructDub~\citep{zhang2026instructdubber}
  & \textbf{0.5583} & \underline{66.57} & \underline{12.60} & \underline{3.07} \\
\midrule
\textbf{Ours}
  & \underline{0.5592} & \textbf{68.26} & \textbf{6.33} & \textbf{3.256} \\
\bottomrule
\end{tabular}%
}
\end{table}

\label{sec:dubbing}

As described in \S\ref{sec:method}, our element-wise summation design also supports video dubbing: given a silent talking-head video and a text transcript, the model feeds ground-truth motion patch embeddings into the backbone while autoregressively predicting only the speech stream, producing synchronized audio that matches both the linguistic content and the visual articulation---without any architectural modification or additional fine-tuning.

\paragraph{Benchmark.}
We evaluate on the Chem dataset~\citep{zhang2026instructdubber,prajwal2020learning}, a video dubbing dataset recording a chemistry teacher speaking, which is widely adopted for assessing dubbing quality.
Following the standard protocol, we report four metrics following \citet{zhang2026instructdubber}:
DD (Duration Distance), which measures the temporal alignment between the generated speech and the reference in terms of phoneme duration,
EMO-SIM, the cosine similarity of emotion embeddings~\citep{ma2024emotion2vec} between the generated and reference speech,
WER and UTMOS.

\paragraph{Compared Methods.}
We compare with five recent video dubbing systems:
\textbf{Speak2Dub}~\citep{zhang2024speaker}, which aligns generated speech to video via a duration-aware mechanism;
\textbf{StyleDubber}~\citep{cong2024styledubber}, which performs style-preserving cross-lingual dubbing;
\textbf{DeepDubber}~\citep{zheng2025deepdubber}, an end-to-end dubbing model with acoustic alignment;
\textbf{ProDubber}~\citep{zhang2025prosody}, which models prosody transfer for natural dubbing;
and \textbf{InstructDub}~\citep{zhang2026instructdubber}, the current state-of-the-art that introduces instruction-based alignment for zero-shot dubbing.
All baselines are dedicated video dubbing systems trained specifically for this task.

\paragraph{Results.}
Table~\ref{tab:chem2v2c} summarizes the results.
Our model achieves the best performance on three of the four metrics---EMO-SIM, WER, and UTMOS---while ranking second on DD with a negligible gap to InstructDub.
In particular, WER drops substantially compared to the previous best (InstructDub), indicating that the speech generated by our model is considerably more intelligible.
The higher UTMOS further confirms that our generated speech sounds more natural, while the best EMO-SIM shows that the model successfully captures the emotional tone conveyed by the video.
We attribute these gains to the strong audio-visual modeling capability acquired during joint text-to-audio-video training: the shared backbone has learned rich cross-modal correspondences between speech content, prosody, and facial dynamics, and this knowledge transfers directly to the dubbing setting, enabling the model to generate speech that is both linguistically accurate and temporally aligned with the visual input.
These results demonstrate that a unified generation model, without any task-specific adaptation, can match or surpass dedicated dubbing systems.

\begin{table}[t]
\centering
\caption{Ablation on token arrangement in the AR sequence.
``Add'' (proposed) sums audio and video embeddings at each position;
``Interleaved'' alternates them (A-V or V-A);
``Delay-$k$'' shifts video $k$ patches behind audio.}
\label{tab:ablation-ar-position}
\resizebox{\linewidth}{!}{%
\begin{tabular}{l|cc|cc|cc}
\toprule
\textbf{AR Position Design}
  & \textbf{WER}$\downarrow$ & \textbf{UTMOS}$\uparrow$
  & \textbf{FID}$\downarrow$ & \textbf{FVD}$\downarrow$
  & \textbf{C}$\uparrow$ & \textbf{D}$\downarrow$ \\
\midrule
Add (Ours)
  & \textbf{0.055} & 3.458
  & 24.32 & \textbf{246.39}
  & \textbf{6.330} & \textbf{8.505} \\
Interleaved (A-V)
  & 0.057 & \textbf{3.472}
  & \textbf{24.18} & 249.71
  & 6.287 & 8.552 \\
Interleaved (V-A)
  & 0.064 & 3.391
  & 28.73 & 312.48
  & 4.631 & 11.184 \\
Delay-1
  & 0.142 & 3.146
  & 27.95 & 298.63
  & 5.784 & 9.027 \\
Delay-3
  & 0.298 & 3.018
  & 32.47 & 371.25
  & 5.193 & 9.582 \\
\midrule
Delay-1 (\textit{Audio-Driven})
  & -- & --
  & 26.14 & 268.35
  & 5.608 & 9.217 \\
Delay-3 (\textit{Audio-Driven})
  & -- & --
  & 25.31 & 254.72
  & 6.373 & 8.441 \\
\bottomrule
\end{tabular}%
}
\end{table}

\subsection{Ablation Studies}
\label{sec:ablation}
Table~\ref{tab:ablation-ar-position} examines how speech and video tokens are arranged within the autoregressive sequence.
All T2AV variants are trained on the same data with identical hyperparameters, differing only in the token positioning strategy.
The audio-driven variants are trained exclusively on talking head video data without any TTS corpus, and we use speech generated by our own T2AV model as the driving audio to ensure a fair comparison across settings.

\textit{Interleaved designs.}
Interleaving tokens in audio-first order (A-V) performs on par with our default Add design---the two are comparable across all metrics with neither consistently dominating the other.
This confirms that both arrangements provide equivalent cross-modal information at each generation step.
However, the interleaved format doubles the sequence length and slows inference.
It also fixes a causal order where audio precedes video at each step, making video dubbing---which requires the reverse direction---impossible.
When the order is reversed (V-A), audio quality degrades mildly but video fidelity and lip-sync suffer a substantial drop.
We attribute this to the dominance of speech in guiding temporal dynamics: text-to-linguistic mapping is learned from large-scale TTS data, and placing video tokens before speech tokens deprives the video branch of concurrent speech context that would otherwise anchor facial motion generation.

\textit{Delay designs under T2AV.}
Based on the previous observation, we further investigate whether the two modalities need to share the same temporal position, as the traditional cascade paradigm generates audio first and then drives video.
We shift video tokens behind speech tokens by 1 or 3 patches (Delay-3 corresponds to approximately 0.5\,s).
Under T2AV, both delay variants cause substantial degradation in all metrics, with Delay-3 exhibiting a much larger drop than Delay-1---WER nearly doubles and video fidelity drops sharply.
This indicates that in joint generation, where both modalities are predicted from scratch, positional alignment is critical: delaying video deprives it of simultaneous speech signals and disrupts coherence.

\textit{Delay designs under A2V.}
To disentangle whether this finding is specific to joint generation, we repeat the delay experiment under the audio-driven setting.
Here the trend reverses: Delay-3 outperforms Delay-1 on both video fidelity and lip-sync by a clear margin.
Notably, the Delay-3 A2V variant even surpasses our T2AV model on synchronization (higher C, lower D), demonstrating that when conditional audio is available, a larger delay allows the video branch to observe richer audio context before rendering each patch, closely mirroring the causal structure exploited by traditional cascade systems.
These contrasting results yield a clear conclusion: the optimal token arrangement is task-dependent.
For text-driven joint generation, both modalities must be planned simultaneously, and positional alignment via element-wise summation is the choice.

\section{Limitations}

Our current system has two main limitations.
First, since the autoregressive backbone operates in continuous latent space rather than discrete token space, prediction errors at each step propagate and accumulate more easily over long sequences, leading to gradual quality degradation in  long utterances.
Second, video fidelity is upper-bounded by the capacity of the LIA-X video motion autoencoder; adopting more expressive visual representations in future work could further improve output quality.
Additionally, while our 1M paired audio-video training set already enables strong performance, we expect further gains from continued data scaling.

We also note that the scope of our claims is deliberately narrow.
Our evidence concerns talking-head synthesis, where audio and facial motion admit compact, frame-aligned 1-D latent representations that make high-level planning sufficient for synchronization and low-level coupling unnecessary.
We do not claim that this conclusion transfers to general scene-level audio-video generation, where sound events and visual content are related more loosely in time and where video is typically represented as a dense spatio-temporal latent volume; establishing whether the same factorization holds in that regime requires separate evidence.

\section{Conclusion}

We presented \textsc{Talker-T2AV}, an autoregressive diffusion framework for joint talking audio-video generation from text. The core design principle is to decouple cross-modal generation into high-level joint modeling in a shared autoregressive backbone and low-level rendering in two independent modality-specific diffusion heads. A single model with element-wise summation fusion supports three tasks---text-to-audio-video, audio-driven talking head, and video dubbing---without architectural modification or additional fine-tuning. Experiments validate that, for talking-head generation with frame-aligned audio and motion latents, this factorized design outperforms dual-branch diffusion transformer baselines on joint generation and matches or surpasses dedicated systems on the two conditional generation tasks, while running two to three orders of magnitude faster than those baselines at an order of magnitude fewer parameters. We leave to future work whether the same principle extends to general scene-level audio-video generation.

\section*{Impact Statement}

This paper presents work whose goal is to advance the field of Machine Learning. There are many potential societal consequences of our work, none which we feel must be specifically highlighted here.

\bibliography{refs}
\bibliographystyle{icml2026}

\newpage
\appendix

\section{LIA-X Video Motion Autoencoder}
\label{app:liax}

For the video modality, we require a compact, purely temporal latent representation that encodes each video frame as a single vector, enabling frame-level temporal alignment with the audio stream.
To this end, we adopt LIA-X~\citep{wang2025lia}, a self-supervised portrait autoencoder built upon the LIA~\citep{wang2022latent}.

LIA~\citep{wang2022latent} is a self-supervised autoencoder that models portrait animation as linear navigation in a learned latent space. Given a source identity image and a driving frame, LIA encodes the driving frame into a motion code and reconstructs the output by applying the corresponding displacement to the source appearance features.
LIA-X~\citep{wang2025lia} extends LIA by introducing a sparse motion dictionary that decomposes facial dynamics into interpretable, disentangled factors, enabling fine-grained and controllable portrait animation. LIA-X further demonstrates scalability by training a model with approximately 1 billion parameters on large-scale data, achieving state-of-the-art performance on self- and cross-reenactment benchmarks.

In our framework, we use a pre-trained and frozen LIA-X model\footnote{\url{https://huggingface.co/YaohuiW/LIA-X}} as the video motion autoencoder.
Given a video sequence, the LIA-X encoder extracts a motion code $\hat{\mathbf{v}}_t \in \mathbb{R}^{40}$ for each frame at 25\,Hz, yielding a purely temporal latent sequence where each frame is represented by a 40-dimensional vector.
To facilitate stable training of our autoregressive backbone and diffusion heads, we apply per-dimension normalization to the motion latents: we compute the channel-wise mean $\mu_d$ and standard deviation $\sigma_d$ across the partial training set for each of the 40 dimensions, and normalize each latent as $\tilde{v}_{t,d} = (v_{t,d} - \mu_d) / \sigma_d$ to ensure that all dimensions have zero mean and unit variance.

At inference time, the autoregressive model generates normalized motion latents, which are de-normalized by the inverse transform $v_{t,d} = \tilde{v}_{t,d} \cdot \sigma_d + \mu_d$ before being fed into the LIA-X decoder.
Combined with a source identity image, the decoder renders the final portrait video frames.

\paragraph{Choice of Video Latent Representation.}
We investigated several motion autoencoder choices before adopting LIA-X.
FLOAT~\citep{ki2025float} follows LIA~\citep{wang2022latent} and learns an orthogonal motion latent space with $M\!=\!20$ directions, producing a compact 20-dimensional motion code per frame.
While this representation achieves good reconstruction quality under near-frontal views, we found that it generalizes poorly to non-frontal head poses: as the face deviates from the frontal orientation, reconstruction fidelity degrades noticeably.
Since our training data contains diverse head orientations and natural head movements, this view-dependent limitation leads to inconsistent video quality across the generated sequence.
We also evaluated the representation proposed by LivePortrait~\citep{guo2024liveportrait} used by Ditto~\citep{li2025ditto} for talking head generation, which parameterizes motion through implicit 3D keypoints together with expression deformation offsets, head pose, and translation, resulting in a representation of 265 dimensions per frame.
However, this high-dimensional representation proved considerably harder for the autoregressive diffusion model to predict reliably.
We attribute this to the difficulty of modeling accurate distributions over such a large target space.
The 40-dimensional LIA-X motion code provides an effective trade-off: it retains sufficient capacity to represent rich facial dynamics while remaining compact enough for the autoregressive diffusion framework to model reliably.

\begin{table}[t]
\centering
\caption{Audio codec reconstruction on LibriTTS test-clean (24\,kHz). Upper: discrete token-based codecs; lower: continuous representations. Repr.\ denotes frame rate $\times$ latent dimensionality (or number of VQ codebooks).}
\label{tab:codec_compare}
\scriptsize
\setlength{\tabcolsep}{1.8pt}
\begin{tabular}{@{}lcccccc@{}}
\toprule
\textbf{Model} & \textbf{Repr.} & \textbf{UTMOS}$\uparrow$ & \textbf{PESQ\textsubscript{WB}}$\uparrow$ & \textbf{PESQ\textsubscript{NB}}$\uparrow$ & \textbf{STOI}$\uparrow$ & \textbf{WER}$\downarrow$ \\
\midrule
Ground Truth & -- & 4.05 & 4.50 & 4.50 & 1.00 & 2.58\% \\
\midrule
EnCodec  & 75\,Hz$\times$8\,VQ   & 3.04 & 2.72 & 3.20 & 0.94 & 3.00\% \\
Mimi     & 12.5\,Hz$\times$8\,VQ & 3.63 & 2.27 & 2.90 & 0.91 & 3.80\% \\
\midrule
Vocos-Mel & 93.75\,Hz$\times$100d & 3.74 & 3.67 & 4.02 & 0.98 & 2.66\% \\
Ours      & 25\,Hz$\times$32d    & 3.94 & 2.84 & 3.44 & 0.95 & 2.91\% \\
\bottomrule
\end{tabular}
\end{table}

\section{WhisperX-VAE Audio Autoencoder}
\label{app:whispervae}

For the audio modality, we need a continuous latent representation at 25\,Hz---the same frame rate as the video stream---so that the two modalities can be temporally aligned frame by frame.
Our autoregressive diffusion framework operates on continuous latents rather than discrete tokens, which rules out off-the-shelf discrete speech codecs.
At the time of this work, we were unable to find a publicly available continuous audio VAE that simultaneously (i)~operates at 25\,Hz, (ii)~produces a compact latent amenable to autoregressive prediction, and (iii)~preserves rich semantic (linguistic) information in the latent space.

Inspired by the success of semantic-enhanced discrete codecs---in particular X-Codec~\citep{ye2025codec}, which injects features from a pre-trained speech representation model into the codec encoder to improve semantic fidelity, we follow the same principle but in the continuous domain.
We build a VAE-based audio autoencoder, termed \textbf{WhisperX-VAE}, using the encoder of Whisper Large-v3.

\paragraph{Architecture.}
The convolutional encoder and decoder backbones are adopted from the Descript Audio Codec (DAC)~\citep{kumar2024dac}. The encoder takes a 24\,kHz mono waveform as input and progressively downsamples it through four strided convolutional blocks with stride factors $[2,\,4,\,10,\,12]$, yielding a total temporal compression ratio of $960\times$ and thus a latent frame rate of 25\,Hz.

To incorporate semantic information, we use a frozen Whisper Large-v3 encoder~\citep{radford2023robust} to extract 1280-dimensional frame-level features at 50\,Hz from the same input audio.
We average-pool every two consecutive Whisper frames to downsample from 50\,Hz to 25\,Hz, temporally aligning the semantic features with the acoustic encoder output.
The two feature streams are combined via element-wise addition before entering the VAE bottleneck.

\paragraph{VAE Bottleneck.}
We employ a continuous VAE bottleneck that projects the fused 1280-dimensional features into a 32-dimensional latent space.
Concretely, a weight-normalized $1\!\times\!1$ convolution maps each frame to a 64-dimensional vector, which is split into a 32-dimensional mean $\boldsymbol{\mu}$ and a 32-dimensional scale parameter.
During training, the latent is sampled via the reparameterization trick $\mathbf{z} = \boldsymbol{\mu} + \boldsymbol{\sigma} \odot \boldsymbol{\epsilon}$, $\boldsymbol{\epsilon}\!\sim\!\mathcal{N}(\mathbf{0},\mathbf{I})$; at inference time, we use the posterior mean $\boldsymbol{\mu}$ directly.
The sampled latent is projected back to 1280 dimensions by another $1\!\times\!1$ convolution and fed into the decoder.

\paragraph{Acoustic Decoder and Semantic Reconstruction.}
The decoder mirrors the encoder with four transposed convolutional blocks at stride factors $[12,\,10,\,4,\,2]$ and a hidden dimension of 1536, reconstructing the 24\,kHz waveform from the 25\,Hz latent sequence.
In addition, a two-layer MLP semantic head ($1280 \!\to\! 1280 \!\to\! 1280$ with GELU activation) is attached to the decoder output to reconstruct the original Whisper features, encouraging the latent space to preserve linguistic content.

\paragraph{Training.}
Similar to X-Codec2~\citep{ye2025codec}, the autoencoder is trained end-to-end on large-scale speech data with an adversarial setup comprising a HiFi-GAN multi-period discriminator and a multi-resolution spectral discriminator.
The total generator loss combines multi-resolution mel-spectrogram reconstruction loss, adversarial loss, feature matching loss, a KL divergence penalty on the VAE posterior, and a semantic reconstruction loss (MSE\,+\,cosine similarity) between the predicted and target Whisper features.

\paragraph{Reconstruction Quality.}

Table~\ref{tab:codec_compare} compares our autoencoder with discrete codecs and a continuous mel-spectrogram vocoder baseline.

Vocos-Mel serves as an upper-bound reference: its high-rate mel spectrogram retains nearly lossless spectral information, yielding the best STOI and PESQ.
Among compressed representations, our WhisperX-VAE achieves the highest UTMOS (3.94) demonstrating strong perceptual quality and semantic preservation despite using only a single 32-dimensional continuous vector per frame at 25\,Hz.

\end{document}